\documentclass[letterpaper, 10 pt, conference]{ieeeconf}  

\IEEEoverridecommandlockouts                              

\usepackage{graphics} 
\usepackage{epstopdf}
\usepackage{epsfig} 
\usepackage{subfigure}
\usepackage{amsmath} 
\usepackage{multirow}
\usepackage{threeparttable}
\usepackage{enumerate}
\usepackage{array}

\title{\LARGE \bf
Appearance-Invariant 6-DoF Visual Localization using Generative Adversarial Networks
}

\author{Yimin Lin, Jianfeng Huang, Shiguo Lian
\thanks{Yimin Lin, Jianfeng Huang, Shiguo Lian, are all with the AI Innovation and Application Center, China Unicomn, Beijing, 100102, China
        {\tt\small {e-mail: linym39, huangjf53, liansg}@chinaunicom.cn}}%
}

\begin{document}

\maketitle
\thispagestyle{empty}
\pagestyle{empty}

\begin{abstract}
We propose a novel visual localization network when outside environment has changed such as different illumination, weather and season. The visual localization network is composed of a feature extraction network and pose regression network. The feature extraction network is made up of an encoder network based on the Generative Adversarial Network CycleGAN, which can capture intrinsic appearance-invariant feature maps from unpaired samples of different weathers and seasons. With such an invariant feature, we use a 6-DoF pose regression network to tackle long-term visual localization in the presence of outdoor illumination, weather and season changes. A variety of challenging datasets for place recognition and localization are used to prove our visual localization network, and the results show that our method outperforms state-of-the-art methods in the scenarios with various environment changes.

\end{abstract}

\section{INTRODUCTION}
Over the past decade, there has been a great interest in robust long-term 6-DoF visual localization, which aims to retrieve the camera pose including position and orientation with respect to a known environment map~\cite{1piasco2018survey, lowry2016visual}. In fact, it is a challenging task especially when the images change in appearance due to variety of illumination, weather, or season over a long time. The key issue is how to extract the appearance-invariant representation used for 6-DoF camera pose regression in such condition~\cite{2kunze2018artificial}.

Traditionally, various types of handcrafted features have been investigated for vision based localization, such as SIFT, SURF and ORB~\cite{rublee2011orb}, etc.  Most of these features work well in matching scenes under similar illumination condition. However, the feature descriptions would always be affected by drastic appearance changes, which leads to poor localization accuracy. With the research of deep learning, pre-trained Convolutional Neural Networks (CNNs) features~\cite{yi2016lift, 3schonberger2017comparative} are gaining importance in most visual localization tasks and turn out to outperform traditional handcrafted features under different illumination or weather conditions. As a learning-based method, the training of CNNs features needs lots of training data composed of paired images at the same scene under different environment conditions, which is time consuming and infeasible. This is an apparent limitation for CNNs features based long-term visual localization.

\begin{figure}[t]
\centering
\includegraphics[width=0.50\textwidth]{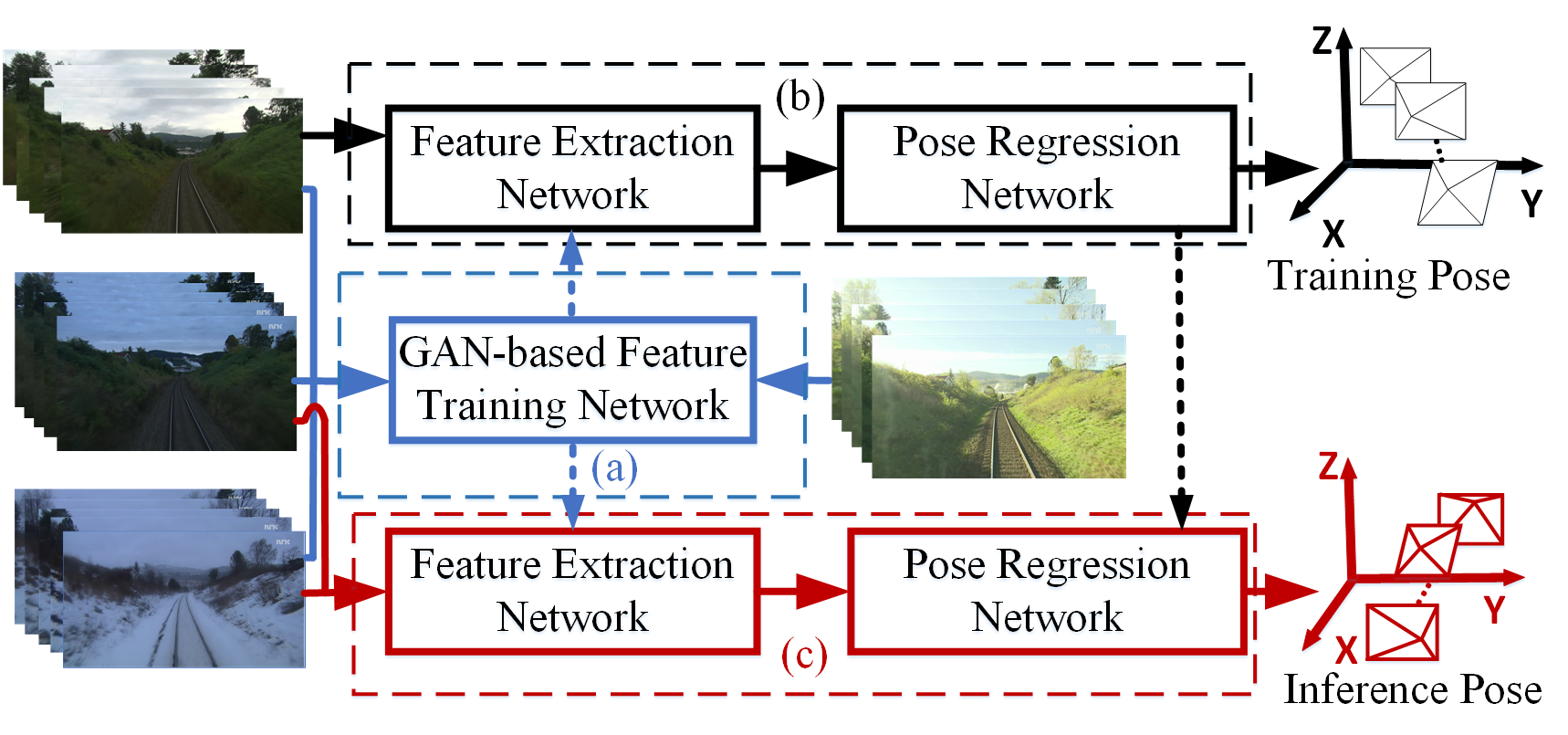}
\caption{A demonstration of our proposed method for long-term 6-DoF visual localization. It is composed of a GAN-based feature training network (a) and feature based localization network (for training (b) and inference (c)). For example, various season data (e.g.,spring, summer, autumn and winter) are used for training the networks.}
\label{figure:figure1}
\end{figure}

To address this issue, recent works exploited domain adaptation techniques to reduce the disturbance of  environment changes. Building on the flexibility of Generative Adversarial Networks (GANs)~\cite{goodfellow2014generative}, recent works model the transformation from a source domain (eg. summer) to a target domain (eg. winter) as the image preprocessing for place recognition. However, they only paid attention to synthesize images of the same location under different conditions and each method could only solve one kind of change. Different from previous methods, our work aims at using the encoder networks of GAN such as CycleGAN~\cite{zhu2017unpaired} to capture intrinsic appearance-invariant feature maps from unpaired samples under different weather and season conditions, as shown in Fig.~\ref{figure:figure1}. In the experimental section, we will show that the feature map focusing on structural information is more stable when illumination, weather or season changes. Moreover, some CNNs have recently been explored to directly regress the 6-DoF metric pose~\cite{kendall2015posenet}. However, they are still unable to outperform state-of-the-art feature-based localization. This is partly due to their inability to internally model the 3D structural constraints of the environment while learning from limited training data. Therefore, our feature extraction network is also helpful to capture invariant feature and improve the performance of 6-DoF pose regression in various scenarios. The main contributions of this paper can be summarized as follows:

1. We design a learning network and strategy based on CycleGAN to capture appearance-invariant feature map from the image captured in different season or weather conditions;

2. We incorporate the feature extraction network and pose regression network to improve the performance of learning-based 6-DoF pose regression;

3. We also conduct extensive experiments on three public benchmarks, and achieve consistently better performance than other state-of-the-art visual place recognition and localization approaches.

\section{RELATED WORK}\label{sec:review}

In this section, we give a brief overview of previous works that aim to address the challenge of long-term visual localization in the presence of outdoor lighting, weather and season changes.

\subsection{2D Image Feature}\label{sec:review:slam}

Bag-of-Word (BOW) approaches~\cite{sivic2003video} have been  proposed in the early years towards developing a system to constitute handcrafted feature descriptors, such as SIFT and ORB, etc. These methods rely entirely on the feature detector to determine what parts of the image are useful for localization, as well as relying on the invariant feature descriptor to provide robustness to changes in appearance. Unfortunately, handcrafted feature matching is not robust to appearance changes due to strong lighting variation, weather conditions and seasonal changes. Therefore, these feature based BOW methods have not achieved state-of-the-art performance in a long-term visual localization. Some approaches also exploit sequential information to alleviate the problem of spatially inconsistent and non-perfect image matches. SeqSLAM~\cite{milford2012seqslam} attempts to match sequences of images rather than individual images. It substantially reduces the false positive matches and recognizes places that underwent heavy changes in appearance (e.g. from summer to winter). Inspired by this work,~\cite{naseer2018robust} generated network flow with the sequential information for higher gain in the long-term localization performance.

\subsection{Learning-based Feature}\label{sec:review:slam}

In recent years, some CNNs have shown tremendous progress in learning better visual descriptors of whole images or image regions to improve place recognition performance. ~\cite{sunderhauf2015performance} presented a method to combine the individual strengths of the high-level and mid-level feature layers to partition the search space and recognize places under severe appearance changes.~\cite{chen2017deep} trained two CNN models to learn condition-invariant features for place recognition across extreme appearance conditions.~\cite{chen2017only} took a step deeper into the internal structure of CNNs and proposed novel CNN-based image features for place recognition by identifying salient regions and creating their regional representations directly from the convolutional layer activations. By training with millions of images from a wide variety of environments, these CNNs obtained descriptors which can learn how appearance changes over time.

\subsection{Semantic Methods}\label{sec:review:slam}

Recently, ~\cite{schonberger2018semantic} found that above methods do not work well in practice since they need an expensive pairwise comparison of descriptors, and proposed the method to learn features from high-level 3D geometric and semantic information, which are geometrically stable over a long period time.~\cite{naseer2017semantics} and~\cite{stenborg2018long} designed a localization algorithm based on semantically segmented images and a semantic point feature map. They show that localization works well when the more robust scene descriptor is learnt from semantic information.

\subsection{Domain Adaptation}\label{sec:review:slam}

Some recent studies considered domain adaptation and transfer learning techniques to address robust localization in such varying environmental conditions. For instance,~\cite{wulfmeier2017addressing} solved the problem in the context of domain adaptation by developing a framework for applying adversarial techniques to align features across domains. ~\cite{mancini2018robust} proposed a novel domain generalization framework to learn classification models that are able to generalize to unseen scenarios.~\cite{porav2018adversarial} effectively compared images under different illumination by visually translating one to another domain and then comparing with CycleGAN~\cite{zhu2017unpaired}. They enforced a descriptor-aiding feature loss on the input and synthetic image in hopes of these features staying present in the intermediate translated image. The main drawback is that they localize in two steps since the network output a synthetic image instead of invariant feature for place retrieval. Similar to our approach,~\cite{latif2018addressing} used a pair of coupled Generative Adversarial Networks (GANs) to generate the appearance of one domain from another without requiring image-to-image correspondences across the domains. Then the feature spaces learned by the discriminator for each domain are more useful for the place recognition task. However, it needs to learn a new model again when the source domain corresponds to another domains. The reason is that they are trained and inferred from only two domains each time, which is more complex than us.  In contrast, we can learn unified feature from different domains as inputs with only once training.

\subsection{Learning-based Visual Relocalization}\label{sec:review:slam}

Learning-based relocalization systems perform as image classification with very large scale datasets. For example, Kendall et al. proposed PoseNet~\cite{kendall2015posenet}, which was the first successful end-to-end pre-trained deep CNNs approach for 6-DoF pose regression. In addition, Clark et al.~\cite{walch2017image} introduced deep CNNs with Long-Short Term Memory (LSTM) units to avoid overfitting to training data. More recently, VLocNet~\cite{valada2018deep} proposed global and Siamese-type relative networks for inferring global poses with the great help of relative motion. Another representative DSAC++~\cite{brachmann2018learning} is a fully convolutional neural network for densely regressing pose from input image to the 3D scene space. To our knowledge, there is no learning-based method reported to tackle the long-term relocalization right now. The reason is that changes in environment, such as weather, illumination, season and so on, will lead to great difference in appearance, which makes it hard for long-term visual relocalization.

\begin{figure*}[t]
\centering
\includegraphics[width=1.0\textwidth]{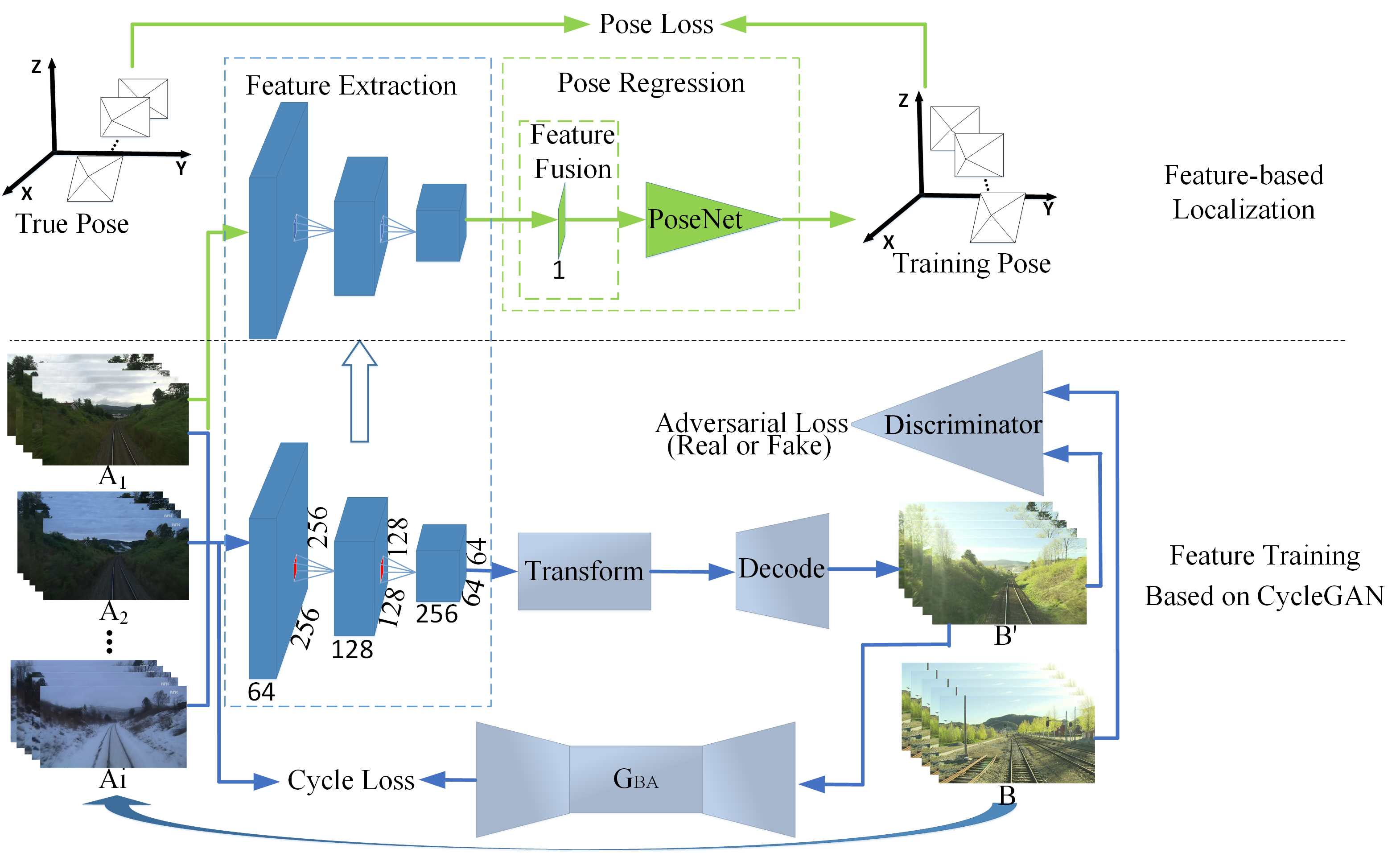}
\caption{The network architecture for appearance-invariance feature extraction and pose regression. The data flow of feature extraction network is described with blue lines while the data flow of pose regression network is described with green lines.}
\label{figure:figure2}
\end{figure*}

\section{OUR APPROACH}\label{sec:review}

As shown in Fig.~\ref{figure:figure2}, our proposed method for long-term visual localization system consists of two stages: feature training based on CycleGAN, and feature-based localization based on the feature extraction network and  pose regression network.

\subsection{Feature Training Based on CycleGAN}\label{sec:review:slam}

Our goal is to maximize the performance of a supervised pose regression task not only in the source domain where labels (ground-truth poses) are available, but also in the unlabeled target domain. Here, we try to use domain transformation network CycleGAN~\cite{zhu2017unpaired}, which learns the image transformation between two different domains without paired samples. And it will be proved that the cycle consistency and general adversarial training strategy are available to capture the invariant distribution of the target domain in the encoding network. As shown in Fig.~\ref{figure:figure2}, ${A_i}(i = 1:M)$ and $B$, where $\{ {a_{ij}}\} (j = 1:N)$, ${a_{ij}} \in \{ {A_i}\}$ and $\{ {b_k}\} (k = 1:S)$, ${b_k} \in B$, are images collected under significantly different lighting, weather and seasonal conditions. Assuming that there exists a generator function ${G_{AB}}$ that can transform domain ${A_i}$ into domain $B$, such that $\widetilde b = {G_{AB}}({a_{ij}})$. Additionally, another generator ${G_{BA}}$ transforms in the reverse direction, that is $\widetilde a = {G_{BA}}(\widetilde b)$. Similarly, ${G_{BA}}$ should also transform domain $B$ to ${A_i}$. So we apply a cycle consistency ${L_1}$ loss between domain ${A_i}$ and $B$ like:

\begin{equation}\label{eq:2}
\begin{array}{l}
{L_{cyc}}({G_{AB}},{G_{BA}}) = {E_{a - {P_{data}}}}_{(a)}[||{G_{BA}}({G_{AB}}(a) - a|{|_1}] + \\
\;\;\;\;\;\;\;\;\;\;\;\;\;\;\;\;\;\;\;\;\;\;\;\;\;\;{E_{b - {P_{data}}}}_{(b)}[||{G_{AB}}({G_{BA}}(b) - b)|{|_1}]
\end{array}
\end{equation}

where $a-{P_{data}}(a)$ and $b-{P_{data}}(b)$ are the collections of image at the correspond domain, and $E$ is the expectation function.
Discriminator $D_A$ and $D_B$ work in each domain and try to discriminate between $A,\widetilde A$ and $B,\widetilde B$ respectively. For simplicity, we only give discriminator $D_A$ in Fig.~\ref{figure:figure2}. And we apply adversarial losses like:

\begin{equation}\label{eq:2}
\begin{array}{l}
{L_{Gan}}({G_{AB}},{D_B}) = {E_{b - {P_{data}}}}_{(b)}[\log {D_B}(b)] + \\
\;\;\;\;\;\;\;\;\;\;\;\;\;\;\;\;\;\;\;\;\;\;\;\;\;\;{E_{a - {P_{data}}(a)}}[1 - \log {D_B}({G_{AB}}(a))]\\
{L_{Gan}}({G_{BA}},{D_A}) = {E_{a - {P_{data}}}}_{(a)}[\log {D_A}(a)] + \\
\;\;\;\;\;\;\;\;\;\;\;\;\;\;\;\;\;\;\;\;\;\;\;\;\;{E_{b - {P_{data}}(b)}}[1 - \log {D_A}({G_{BA}}(b))]
\end{array}
\end{equation}

Then the task of image-to-image transformation GAN is to minimax cycle and adversarial loss as:

\begin{equation}\label{eq:2}
\begin{array}{l}
\mathop {\min }\limits_G \mathop {\max }\limits_D \{ {L_{Gan}}({G_{AB}},{D_B})\}  + \\
{L_{Gan}}({G_{BA}},{D_A}) + \omega {L_{cyc}}({G_{AB}},{G_{BA}})\}
\end{array}
\end{equation}

\begin{figure}[h]
\centering
\includegraphics[width=0.45\textwidth]{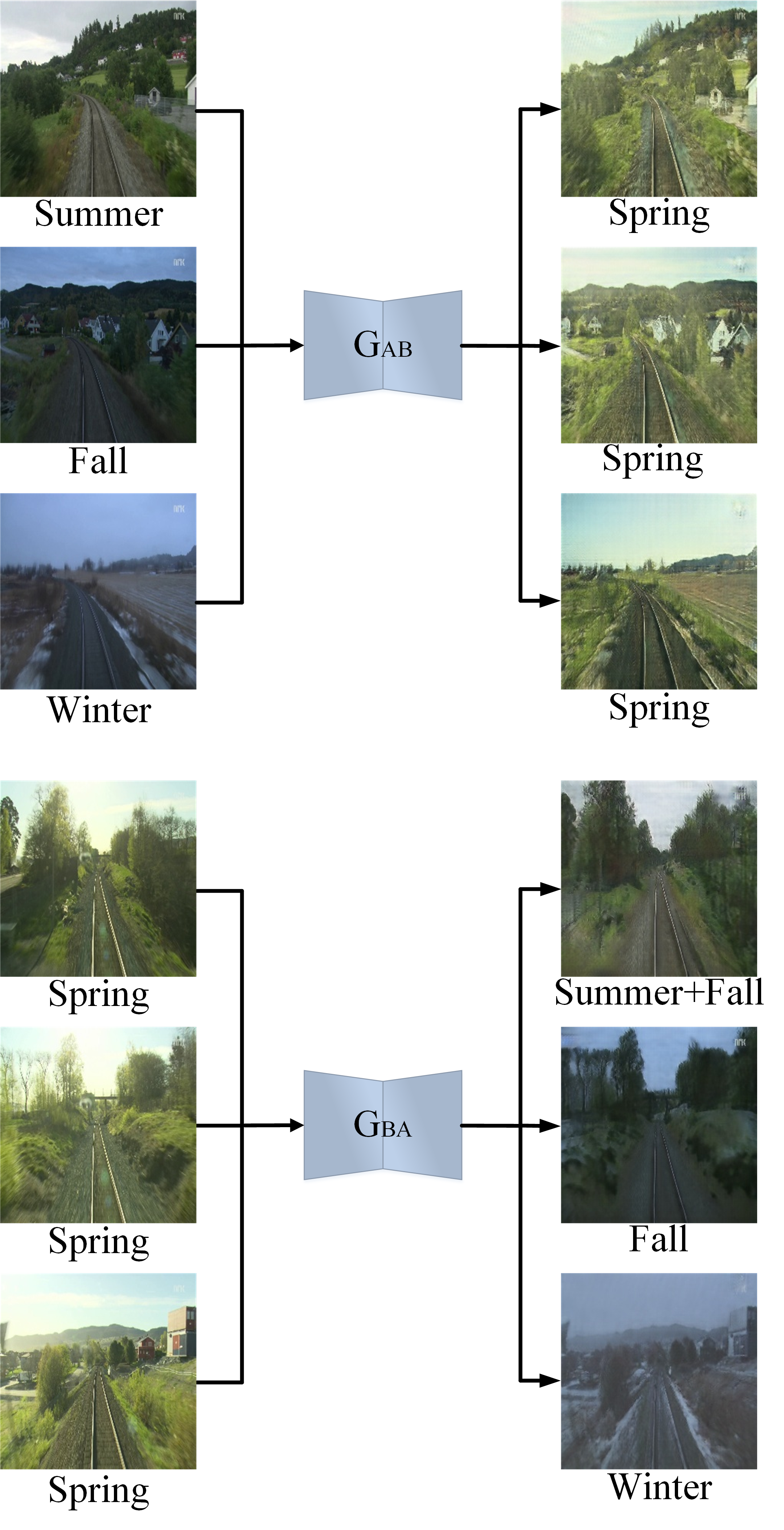}
\caption{Domain transformation results of generator $G_{AB}$ and $G_{BA}$.}
\label{figure:figure30}
\end{figure}

where $\omega$ is a weight to balance the training direction. Here we wish to find stable feature representation in domain $A_i$, where images are collected under significantly different lighting, weather and seasonal conditions. For instance, when we want to find the appearance-invariant feature maps of images taken in summer, fall and winter, therefore, we divide the images captured in summer, fall and winter into domain $A$, denoted as ${A_1},{A_2}$ and ${A_3}$ respectively. Naturally, spring images are divided into domain $B$ as Fig.~\ref{figure:figure2} shows. And Fig.~\ref{figure:figure30} shows domain transformation results of generator $G_{AB}$ and $G_{BA}$.

From Fig.~\ref{figure:figure30}, we can see that generator $G_{AB}$ transfers summer, fall and winter image into spring image. Different from $G_{AB}$, the transformation result of $G_{BA}$ may be the combination of multi-season or a single season, which is determined by the content of the input image.

\subsection{Feature-based Localization}\label{sec:review:slam}

The proposed feature-based localization is composed of feature extraction network and pose regression network. The feature extraction network corresponds to the encoding network of generator $G_{AB}$ based on CycleGAN, and each layer of the feature extraction network outputs a tensor sized $a*b*M$, where $a*b$ is the size of feature map while $M$ means the number of channel. For each layer, we obtain fusion future by fusing $M$ feature maps as follow:

\begin{equation}\label{eq:2}
{f_{fus}} = \sum\limits_{i = 1}^M {{f_i}}
\end{equation}

where $f_{fus}$ is the fusion feature and $f_i$ means the ith feature map.

By feeding fusion feature into a pose estimation network, eg. PoseNet, we can estimate a 6-dimensional pose $P = \{ {p_i}\}$ with ${p_i} \in SE(3)$ . Similar to previous works~\cite{kendall2015posenet}, we regress the pose on Euclidean loss using stochastic gradient descent with the following objective loss function:
\begin{equation}\label{eq:2}
loss(I) = ||\widehat x - x|{|_2} + \beta ||\widehat q - \frac{q}{{||q||}}|{|_2}
\end{equation}

where $\beta$ is a scale factor to balance the position and orientation errors.

\begin{figure}[h]
\centering
\includegraphics[width=0.45\textwidth]{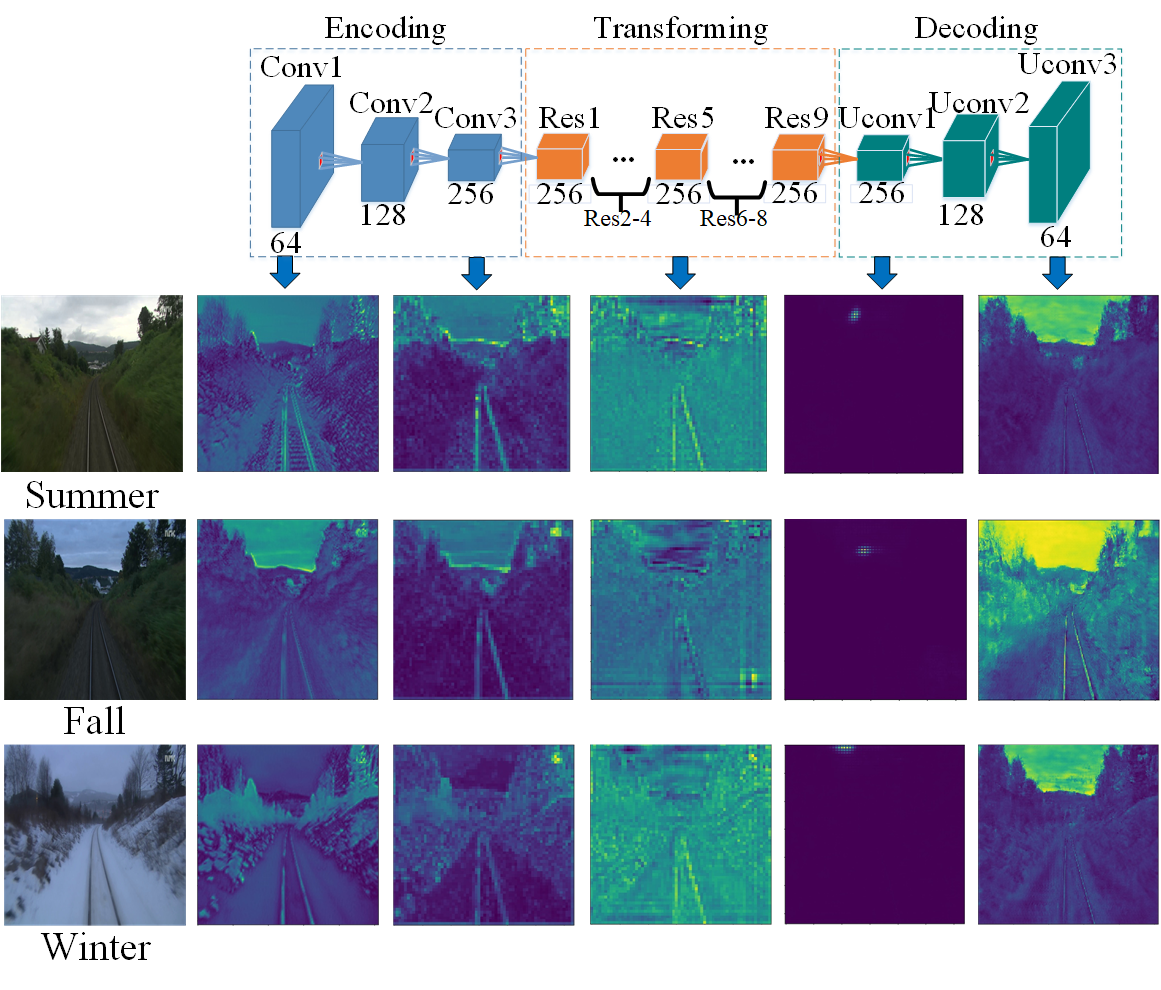}
\caption{Architecture of $G_{AB}$ and fusion features from five layers. Top row shows fifteen network layers of $G_{AB}$, terms on the top of network layers represent operation, eg. Conv1 means the first convolutional layer, Res1 means the first residual block and Uconv1 means the first unconvolutional layer, and number on the bottom of layers correspond to the channels of the output from this layer. The images in 1st column are input images and the rest columns are fusion features from different layers of $G_{AB}$.}
\label{figure:figure3}
\end{figure}

Actually, fusion feature from different layer has different robustness against appearance changes, therefore, we need to find the most stable fusion feature. In implementation, the feature extraction network $G_{AB}$ includes fifteen layers with their sizes shown in Table~\ref{table1}. With three scenes (summer, fall, winter) for example, we computer each layer's fusion feature with Eq. (4). Fig.~\ref{figure:figure3} shows five fusion features for each scene to visualize what our network learns. It can be seen that the lower-level fusion feature from Conv1 typically varies significantly under different seasons. While the higher-level fusion feature from Conv3 focuses mostly on geometric information and is comparatively invariant.

\begin{table}[!htbp]
	\centering
	\caption{The layers of $G_{AB}$ and their output dimensionality.}
	\label{table1}
	\begin{tabular}{cccc}
		\hline
        Layer&Dimensions&Layer&Dimensions\\
		\hline
         Conv1&256x256x64&Uconv1&64x64x256\\
         Conv2&128x128x128&Uconv2&128x128x128\\
         Conv3&64x64x256&Uconv3&256x256x3\\
         Res1-9&64x64x256&&\\
        \hline
	\end{tabular}
\end{table}

 \begin{figure}
\centering
\includegraphics[width=0.45\textwidth]{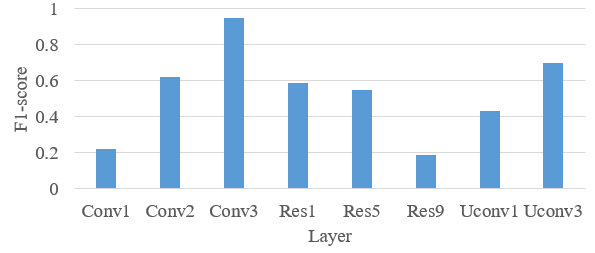}
\caption{F1-score of different layer on Nordland dataset. 15 layers' F1-scores are computed based on summer and fall image sets each composed of 400 images, with only 8 listed here.}
\label{figure:figure5}
\end{figure}

Furthermore, we use the F1-score to quantitatively find the layer robust against environment changes. Firstly, we select paired summer and fall image sets, each of which composed of 400 images. Then, we obtain 15 fusion features (each for one layer) for each image in the two image sets using Eq. (4). Finally, we compute each layer's F1-score by analyzing the similarity between summer images' features and fall images' features, which is shown in Fig.~\ref{figure:figure5}. It is easy to find that Conv3 layer has the maximal F1-score, which means that fusion feature from Conv3 layer is appearance-invariant across season. Therefore, we select the fusion feature from Conv3 layer as our appearance-invariant feature.

\section{Experimental Results}\label{sec:review}

In this section, we firstly give a brief introduction on the challenge datasets, and then describe the training details of our feature extraction and localization models. At last we evaluate the robustness of our appearance-invariant features in cross-time place recognition and visual localization.

\subsection{Datasets}\label{sec:review:slam}

\textbf{Nordland} dataset~\cite{sunderhauf2013we} has been recorded from the perspective of the train over four-different seasons: summer, winter, fall and spring. It consists of about 10 hours of video under each weather condition. Nordland is regarded as a perfect experimentation dataset to test place recognition algorithms on pure appearance changes.

\textbf{Oxford RobotCar} dataset~\cite{maddern20171} contains almost 20 million images collected from 6 cameras mounted to the vehicle. By frequently traversing the same route over the period of a year, Oxford Robotcar datasets have made a great contribution to long-term localization in real-world, dynamic urban environments.

\textbf{Virtual KITTI} dataset~\cite{gaidon2016virtual} generates large, photo realistic, varied datasets of synthetic videos (21,260 frames), automatically and densely labeled for various video understanding tasks. These videos were created using the Unity game engine and a novel real-to-virtual cloning method. This allows for the quantitative study of special effects to simulate different lighting and weather conditions in our case.

\subsection{Implementation Details}\label{sec:review:slam}

\begin{figure}[h]
\centering
\includegraphics[width=0.5\textwidth]{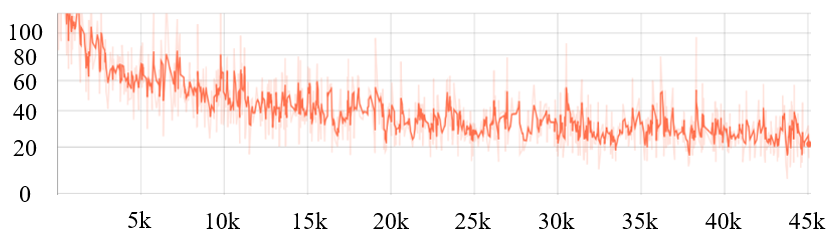}
\caption{Training loss of feature extraction. X axis corresponds to iterations while Y axis corresponds to the loss defined in Eq. (3).}
\label{figure:figure4}
\end{figure}

Our experiments run on a desktop with an Intel core i7-6700K CPU @4.00GHz$\times$8, and all the models are trained with Tensorflow deep learning framework on a NVIDIA GeForce GTX 1070. Our network includes two parts: feature extraction model and pose regression model. Feature extraction network is used to obtain the stable features, which then used as the inputs of place recognition and visual localization tasks under illumination variations or different season conditions. For higher training speed and better results, we use instance normalization after convolution operation when training our feature extraction model. At the same time, Relu is used to activate these layers. We design our feature extraction loss as Eq. (3) with $\omega {\rm{ = }}20$, which contains the loss of image domain A to domain B, domain B to domain A and the cycle loss. Meanwhile, we select Adam solver as the optimization algorithm with ${\beta _1}{\rm{ = }}0.9,{\kern 1pt} {\beta _2}{\rm{ = }}0.999{\kern 1pt}$ and $e = {10^{ - 10}}$. For all experiments, we train our feature extraction network using a fixed learning rate 0.0002, and the batch size is set to 1. The image fed into our feature extraction network is resized to 256$\times$256. We set a minimum loss, and Fig.~\ref{figure:figure4} shows the loss curve as a function of iterations when training our network. We can see that the loss converges at 20 after about 45000 iterations, which costs about 48 hours.

After training the feature extraction network, we freeze the weights and feed its output into a pose regression network. The pose regression network based on GoogLeNet, whose input is image with 256$\times$256 resolution and output is the corresponding pose. However, the feature extraction network outputs feature map with 64$\times$64 resolution, which is different from 256$\times$256. To solve this problem, we resize the feature map to 256$\times$256. We train our localization network with a fixed learning rate of 0.0001 with a batch size of 75. To accelerate convergence, we initialize the weights with a Gauss function, and set a maximum iteration of 15000.

\subsection{Cross-Time Place Recognition}\label{sec:review:slam}

 Our appearance invariant feature can be used for place recognition when the environment changes, such as weather, illumination and season variation. Experiments on Nordland and VKITTI dataset are used to evaluate our feature extraction network when season and illumination change. When training feature extraction model, we divide the four seasons in Nordland dataset into two domains. One domain consists of winter, summer and fall, and the other is made up of spring. The training set of VKITTI consists of morning, rain and overcast images in the sixth scene. Similar to Nordland dataset, we divide the images into two domains, one consists of morning and overcast images while the other contains rain images. There are total nine groups of experiments for place recognition, and each includes 400 query images and 400 database images.

 SeqSLAM~\cite{milford2012seqslam}, NetVLAD~\cite{arandjelovic2016netvlad} and Naseer et al.~\cite{naseer2018robust} are chosen to compare with our feature extraction network in place recognition task. SeqSLAM is a place recognition algorithm based on image sequence, which can recognize the same scene when season and illumination change. We select open SeqSLAM to compare with our network. By combining normal CNN network with VLAD layer, NetVLAD realizes an end-to-end place recognition. We select high performance NetVLAD combination (VGG-16, NetVLAD with whitening) as compared method. Naseer et al.~\cite{naseer2018robust} proposed a HOG feature based data association method with network flow for place recognition, and we use the open code on github for test.

\begin{figure}[h]
\centering
\includegraphics[width=0.5\textwidth]{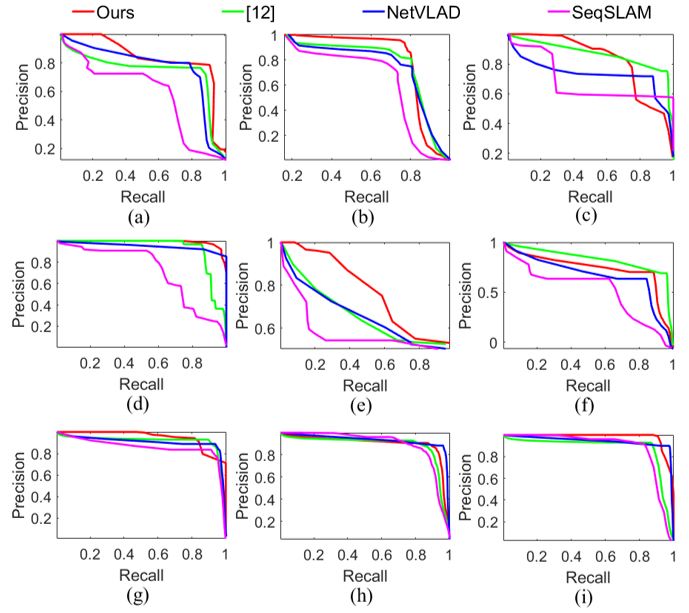}
\caption{ PR-curve of four place recognition networks. X axis represents recall value and Y axis represents precision value. (a)-(f) represent results on Nordland dataset, which respectively correspond to Spring-Summer, Spring-Fall, Spring-Winter, Summer-Fall, Summer-Winter, Fall-Winter scenes. (g)-(i) represent results on VKITTI dataset, which respectively correspond to Morning-Overcast, Morning-Rain, Overcast-Rain scenes.}
\label{figure:figure6}
\end{figure}

We extract the stable feature of all query and database image with our feature extraction network, and then compute the similarity score between query image and database image according to the SSIM of their stable features. By doing this, we can get the similarity score between any two query and database images, which can be described as a 400$\times$400 score matrix. The other three advanced methods can also get a matching score matrix in the same way. By setting different threshold on score matrix, we can compute the precision and recall value. Here, we provide the P-R curves of these four methods in Fig.~\ref{figure:figure6}.

 We can visually observe the performance of these four methods with P-R curve. P-R curve is always used for evaluate classification algorithms, whose x axis and y axis respectively correspond to recall and precision value. The closer to (1, 1) the curve approaches, the better matching result the method has. It is easily to find our P-R is closer to (1,1) on most scenes, prove that our network can exactly recognize the same scene when season or illumination changes.

We also compute the F1-score of the P-R curve for a quantitative comparison. When computing F1-score, we set the weights of recall and precision value to 1. The F1-score results are shown in Table~\ref{table2}. We can know that F1-scores on VKITTI dataset are higher than Nordland dataset. This is because Nordland images are shot in real scene and it is impossible to keep the same scene invariant at different time. However, VKITTI images are generated artificially and the content of image does not change too much. Besides, when experiments include winter images, the F1-score will decrease. The reason is that there is snow in winter scene, which leads to great changes in appearance.

 \begin{table}[!htbp]
	\centering
	\caption{F1-score of the four place recognition networks.}
	\label{table2}
	\begin{tabular}{p{0.6cm}ccccc}
		\hline
        Dataset&Scene&Ours&Naseer~\cite{naseer2018robust}&NetVLAD&SeqSLAM\\
		\hline
        \multirow{6}{*}{Nordland}&Spring-Summer&\textbf{0.84}&0.83&0.82&0.75\\
         &Spring-Fall&\textbf{0.86}&0.84&0.83&0.74\\
         &Spring-Winter&0.76&\textbf{0.80}&0.74&0.73\\
         &Summer-Fall&\textbf{0.95}&0.91&0.90&0.78\\
         &Summer-Winter&\textbf{0.73}&0.69&0.68&0.62\\
         &Fall-Winter&0.74&\textbf{0.75}&0.70&0.63\\
         \hline
         &Average&\textbf{0.81}&0.80&0.78&0.71\\
        \hline
         \multirow{3}{*}{VKITTI}&morning-overcast&0.88&\textbf{0.90}&0.86&0.85\\
         &Morning-Rain&0.89&0.87&\textbf{0.90}&0.84\\
         &Rain-Overcast&\textbf{0.95}&0.89&0.93&0.86\\
         \hline
         &Average&\textbf{0.91}&0.89&0.90&0.85\\
        \hline
	\end{tabular}
\end{table}

On Nordland dataset, our network has a great promotion up to 0.1 compared with SeqSLAM network. NetVLAD has a good result, however, its results are still about 0.03 lower than ours. Naseer et al.~\cite{naseer2018robust} uses image sequence information for place recognition, which makes it less affected across season. Therefore, it has the best performance on spring-winter and fall-winter experiments. However, on the rest four groups of experiments, our network has better result compared with Naseer et al.~\cite{naseer2018robust}. On VKITTI dataset, all methods have a good result expect SeqSLAM. In summary, our network is slightly better than NetVLAD and Naseer et al.~\cite{naseer2018robust}. At the same time, our results are much better than SeqSLAM. The experimental results prove that the feature extracted by our network is robust to season and illumination variation, which can promote the development of place recognition.

\begin{figure}[h]
\centering
\includegraphics[width=0.5\textwidth]{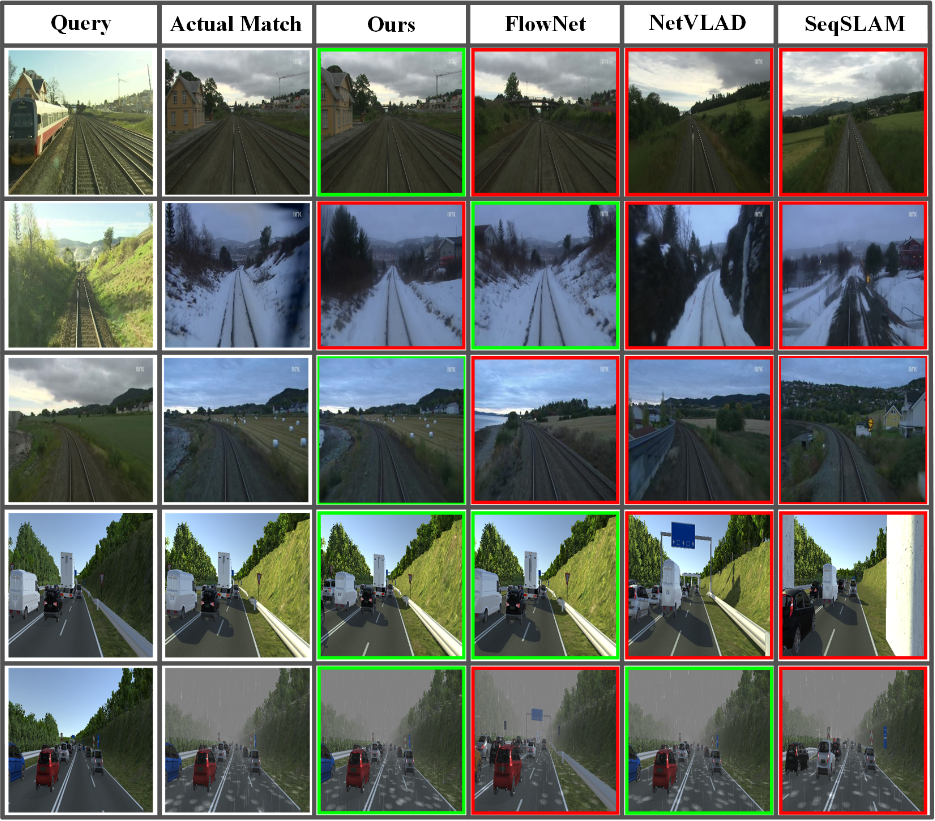}
\caption{Five challenging matching results on Nordland and VKITTI. Each row contains query image (1st column), actual matching image(2nd column) and matching result of different place recognition network (3rd, 4th, 5th and 6th column). The green and red border around result indicate correct and incorrect matching, respectively.}
\label{figure:figure7}
\end{figure}

 For some challenging images, we show the matching result of the four methods in Fig.~\ref{figure:figure7}.  Query images in Fig.~\ref{figure:figure7} usually include two kinds of situations: one is that there are many similar images in the database, the other is that the content of the scene changes. Fig.~\ref{figure:figure7} includes five challenge query images, and our network has the right matching results except the 2nd rows. At second example, there are several images in database similar to query image. Furthermore, there is snow in the database scene, which makes it harder to find the best matching image. Even so, our network can accurately find the matching image at other four examples, which is hard for the other three networks.

\subsection{Cross-Time Localization}\label{sec:review:slam}

In this section, we design the experiment to evaluate our feature-based localization network across season, and then compare our results with the most frequently used pose estimation network PoseNet~\cite{kendall2015posenet}. The comparison experiments run on VKITTI and Oxford Robotcar dataset. VKITTI dataset provides paired images and absolute pose, therefore, we directly use overcast images as training set and morning images as test set. Different from VKITTI dataset, Oxford Robotcar images are non-aligned. To validate our network has exact localization result across season, we select the images shot in February and May as training and test set separately. Then, we need to find the common overlap part of the two groups of images, and then align them. Traditional PoseNet is designed to compute the absolute pose in the scene without considering environment variety. In fairness, we add a group of experiment named PoseNet++, whose training set is composed of images under different environment condtions. We evaluate visual localization model with the mean value of translation error and orientation error. Table~\ref{table3} shows the localization results on VKITTI and Oxford Robotcar datasets, each column includes mean translation(m) and orientation($^\circ$) error.

\begin{table}[!htbp]
	\centering
	\caption{Results of three visual localization networks on VKITTI and Oxford Robotcar with environment changes. Numbers show the mean translation(meter) and orientation(degree) error.}
	\label{table3}
	\begin{tabular}{cccc}
		\hline
        Dataset&PoseNet&PoseNet++&Ours\\
		\hline
         Robotcar&28.6m,$1.8^\circ$&19.3m,$1.1^\circ$&\textbf{10.5}m,$\textbf{0.8}^\circ$\\
         VKITTI&42.2m,$2.5^\circ$&22.1m,$2.0^\circ$&\textbf{12.7}m,$\textbf{1.3}^\circ$\\
         \hline
         Average&35.4m,$2.2^\circ$&20.7m,$1.6^\circ$&\textbf{11.6}m,$\textbf{1.1}^\circ$\\
        \hline
	\end{tabular}
\end{table}

It is easy to find that our localization accuracy on Oxford Robotcar dataset is higher than on VKITTI dataset, the reason is that motion speed between adjacent images on VKITTI dataset is faster, which leads to a larger error. In brief, our network gets the most exact results, while traditional PoseNet has a poorest performance.  Compared with PoseNet, our network has a promotion of 67\% in translation and 50\% in orientation. PoseNet++ gets a better result than PoseNet, even so, the localization error of PoseNet++ is up to 20m in translation, which is about twice bigger than ours. In a word, our model has a great improvement compared with the other two models. Reason for this result is that varieties in illumination and season make the training and test images have big differences in appearance. However, their stable features are similar to each other because they are taken in same scene. Our pose estimation network computes pose with stable features while the other models feed the network with RGB images. In summary, our localization network can estimate pose across season and illumination changes, which brings a great progress for visual localization. We also draw their trajectories as Fig.8 shows. Due to Oxford
Robotcar images are non-aligned, we only circle the overlap trajectory between February and May. It is easy to see that our trajectory is closer to ground truth trajectory, which further proves that our network has accurate localization results.

\begin{figure}[h]
\centering
\includegraphics[width=0.5\textwidth]{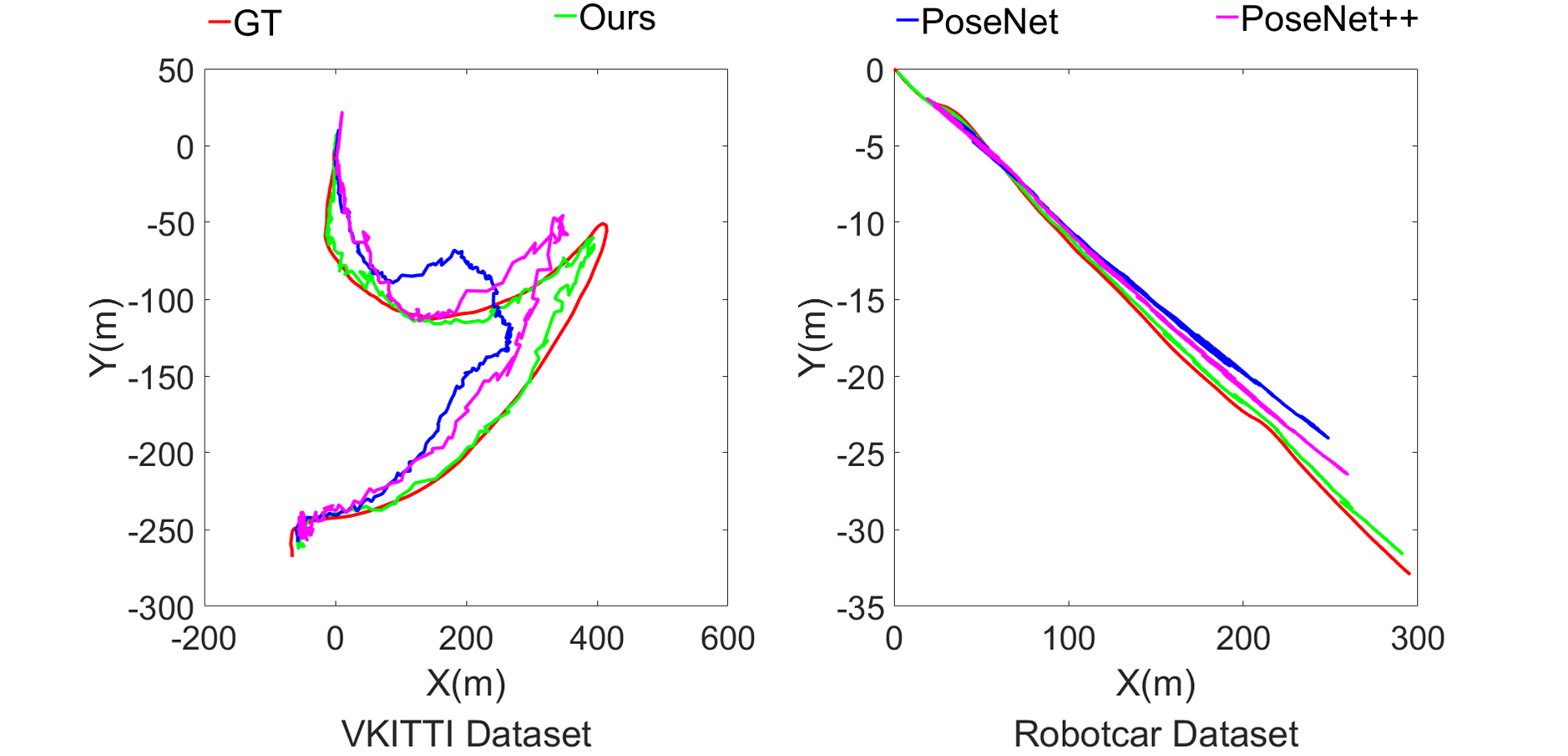}
\caption{Trajectories of three visual localization networks and ground truth(GT) on VKITTI and Robotcar datasets.}
\label{figure:figure8}
\end{figure}

\section{CONCLUSION}\label{sec:review}

In this paper, we pay close attention to solve place recognition and visual localization problems when environment changes a lot. An end-to-end visual localization network is proposed, which includes a feature extraction subnetwork extracting features robust against appearance changes caused by weather, illumination and season changes, and a pose regression subnetwork computing the 6-DoF pose based on such network as PoseNet. The feature extraction subnetwork is obtained by training the CycleGAN with various image sets captured in different conditions. Additionally, the extracted features can be used for place recognition. Experiments on public datasets demonstrate that our network has a better performance compared with the state-of-the-art pose estimation and place recognition networks in the condition of environment changes. Considering that environment change is inevitable in daily life, our work will be important for practical applications.

\bibliographystyle{IEEEtran}
\bibliography{IEEEexample}

\begin{thebibliography}{10}
\providecommand{\url}[1]{#1}
\csname url@rmstyle\endcsname
\providecommand{\newblock}{\relax}
\providecommand{\bibinfo}[2]{#2}
\providecommand\BIBentrySTDinterwordspacing{\spaceskip=0pt\relax}
\providecommand\BIBentryALTinterwordstretchfactor{4}
\providecommand\BIBentryALTinterwordspacing{\spaceskip=\fontdimen2\font plus
\BIBentryALTinterwordstretchfactor\fontdimen3\font minus
  \fontdimen4\font\relax}
\providecommand\BIBforeignlanguage[2]{{%
\expandafter\ifx\csname l@#1\endcsname\relax
\typeout{** WARNING: IEEEtran.bst: No hyphenation pattern has been}%
\typeout{** loaded for the language `#1'. Using the pattern for}%
\typeout{** the default language instead.}%
\else
\language=\csname l@#1\endcsname
\fi
#2}}

\bibitem{1piasco2018survey}
N.~Piasco, D.~Sidib{\'e}, C.~Demonceaux, and V.~Gouet-Brunet, ``A survey on
  visual-based localization: On the benefit of heterogeneous data,''
  \emph{Pattern Recognition}, vol.~74, pp. 90--109, 2018.

\bibitem{lowry2016visual}
S.~Lowry, N.~S{\"u}nderhauf, P.~Newman, J.~J. Leonard, D.~Cox, P.~Corke, and
  M.~J. Milford, ``Visual place recognition: A survey,'' \emph{IEEE
  Transactions on Robotics}, vol.~32, no.~1, pp. 1--19, 2016.

\bibitem{2kunze2018artificial}
L.~Kunze, N.~Hawes, T.~Duckett, M.~Hanheide, and T.~Krajn{\'\i}k, ``Artificial
  intelligence for long-term robot autonomy: A survey,'' \emph{IEEE Robotics
  and Automation Letters}, vol.~3, no.~4, pp. 4023--4030, 2018.

\bibitem{rublee2011orb}
E.~Rublee, V.~Rabaud, K.~Konolige, and G.~Bradski, ``Orb: An efficient
  alternative to sift or surf,'' 2011.

\bibitem{yi2016lift}
K.~M. Yi, E.~Trulls, V.~Lepetit, and P.~Fua, ``Lift: Learned invariant feature
  transform,'' in \emph{European Conference on Computer Vision}.\hskip 1em plus
  0.5em minus 0.4em\relax Springer, 2016, pp. 467--483.

\bibitem{3schonberger2017comparative}
J.~L. Schonberger, H.~Hardmeier, T.~Sattler, and M.~Pollefeys, ``Comparative
  evaluation of hand-crafted and learned local features,'' in \emph{Proceedings
  of the IEEE Conference on Computer Vision and Pattern Recognition}, 2017, pp.
  1482--1491.

\bibitem{goodfellow2014generative}
I.~Goodfellow, J.~Pouget-Abadie, M.~Mirza, B.~Xu, D.~Warde-Farley, S.~Ozair,
  A.~Courville, and Y.~Bengio, ``Generative adversarial nets,'' in
  \emph{Advances in neural information processing systems}, 2014, pp.
  2672--2680.

\bibitem{zhu2017unpaired}
J.-Y. Zhu, T.~Park, P.~Isola, and A.~A. Efros, ``Unpaired image-to-image
  translation using cycle-consistent adversarial networks,'' in
  \emph{Proceedings of the IEEE International Conference on Computer Vision},
  2017, pp. 2223--2232.

\bibitem{kendall2015posenet}
A.~Kendall, M.~Grimes, and R.~Cipolla, ``Posenet: A convolutional network for
  real-time 6-dof camera relocalization,'' in \emph{Proceedings of the IEEE
  international conference on computer vision}, 2015, pp. 2938--2946.

\bibitem{sivic2003video}
J.~Sivic and A.~Zisserman, ``Video google: A text retrieval approach to object
  matching in videos,'' in \emph{null}.\hskip 1em plus 0.5em minus 0.4em\relax
  IEEE, 2003, p. 1470.

\bibitem{milford2012seqslam}
M.~J. Milford and G.~F. Wyeth, ``Seqslam: Visual route-based navigation for
  sunny summer days and stormy winter nights,'' in \emph{2012 IEEE
  International Conference on Robotics and Automation}.\hskip 1em plus 0.5em
  minus 0.4em\relax IEEE, 2012, pp. 1643--1649.

\bibitem{naseer2018robust}
T.~Naseer, W.~Burgard, and C.~Stachniss, ``Robust visual localization across
  seasons,'' \emph{IEEE Transactions on Robotics}, vol.~34, no.~2, pp.
  289--302, 2018.

\bibitem{sunderhauf2015performance}
N.~S{\"u}nderhauf, F.~Dayoub, S.~Shirazi, B.~Upcroft, and M.~Milford, ``On the
  performance of convnet features for place recognition,'' \emph{arXiv preprint
  arXiv:1501.04158}, 2015.

\bibitem{chen2017deep}
Z.~Chen, A.~Jacobson, N.~S{\"u}nderhauf, B.~Upcroft, L.~Liu, C.~Shen, I.~Reid,
  and M.~Milford, ``Deep learning features at scale for visual place
  recognition,'' in \emph{2017 IEEE International Conference on Robotics and
  Automation (ICRA)}.\hskip 1em plus 0.5em minus 0.4em\relax IEEE, 2017, pp.
  3223--3230.

\bibitem{chen2017only}
Z.~Chen, F.~Maffra, I.~Sa, and M.~Chli, ``Only look once, mining distinctive
  landmarks from convnet for visual place recognition,'' in \emph{2017 IEEE/RSJ
  International Conference on Intelligent Robots and Systems (IROS)}.\hskip 1em
  plus 0.5em minus 0.4em\relax IEEE, 2017, pp. 9--16.

\bibitem{schonberger2018semantic}
J.~L. Sch{\"o}nberger, M.~Pollefeys, A.~Geiger, and T.~Sattler, ``Semantic
  visual localization,'' in \emph{Proceedings of the IEEE Conference on
  Computer Vision and Pattern Recognition}, 2018, pp. 6896--6906.

\bibitem{naseer2017semantics}
T.~Naseer, G.~L. Oliveira, T.~Brox, and W.~Burgard, ``Semantics-aware visual
  localization under challenging perceptual conditions,'' in \emph{2017 IEEE
  International Conference on Robotics and Automation (ICRA)}.\hskip 1em plus
  0.5em minus 0.4em\relax IEEE, 2017, pp. 2614--2620.

\bibitem{stenborg2018long}
E.~Stenborg, C.~Toft, and L.~Hammarstrand, ``Long-term visual localization
  using semantically segmented images,'' in \emph{2018 IEEE International
  Conference on Robotics and Automation (ICRA)}.\hskip 1em plus 0.5em minus
  0.4em\relax IEEE, 2018, pp. 6484--6490.

\bibitem{wulfmeier2017addressing}
M.~Wulfmeier, A.~Bewley, and I.~Posner, ``Addressing appearance change in
  outdoor robotics with adversarial domain adaptation,'' in \emph{2017 IEEE/RSJ
  International Conference on Intelligent Robots and Systems (IROS)}.\hskip 1em
  plus 0.5em minus 0.4em\relax IEEE, 2017, pp. 1551--1558.

\bibitem{mancini2018robust}
M.~Mancini, S.~R. Bul{\`o}, B.~Caputo, and E.~Ricci, ``Robust place
  categorization with deep domain generalization,'' \emph{IEEE Robotics and
  Automation Letters}, vol.~3, no.~3, pp. 2093--2100, 2018.

\bibitem{porav2018adversarial}
H.~Porav, W.~Maddern, and P.~Newman, ``Adversarial training for adverse
  conditions: Robust metric localisation using appearance transfer,'' in
  \emph{2018 IEEE International Conference on Robotics and Automation
  (ICRA)}.\hskip 1em plus 0.5em minus 0.4em\relax IEEE, 2018, pp. 1011--1018.

\bibitem{latif2018addressing}
Y.~Latif, R.~Garg, M.~Milford, and I.~Reid, ``Addressing challenging place
  recognition tasks using generative adversarial networks,'' in \emph{2018 IEEE
  International Conference on Robotics and Automation (ICRA)}.\hskip 1em plus
  0.5em minus 0.4em\relax IEEE, 2018, pp. 2349--2355.

\bibitem{walch2017image}
F.~Walch, C.~Hazirbas, L.~Leal-Taixe, T.~Sattler, S.~Hilsenbeck, and
  D.~Cremers, ``Image-based localization using lstms for structured feature
  correlation,'' in \emph{Proceedings of the IEEE International Conference on
  Computer Vision}, 2017, pp. 627--637.

\bibitem{valada2018deep}
A.~Valada, N.~Radwan, and W.~Burgard, ``Deep auxiliary learning for visual
  localization and odometry,'' in \emph{2018 IEEE International Conference on
  Robotics and Automation (ICRA)}.\hskip 1em plus 0.5em minus 0.4em\relax IEEE,
  2018, pp. 6939--6946.

\bibitem{brachmann2018learning}
E.~Brachmann and C.~Rother, ``Learning less is more-6d camera localization via
  3d surface regression,'' in \emph{Proceedings of the IEEE Conference on
  Computer Vision and Pattern Recognition}, 2018, pp. 4654--4662.

\bibitem{sunderhauf2013we}
N.~S{\"u}nderhauf, P.~Neubert, and P.~Protzel, ``Are we there yet? challenging
  seqslam on a 3000 km journey across all four seasons,'' in \emph{Proc. of
  Workshop on Long-Term Autonomy, IEEE International Conference on Robotics and
  Automation (ICRA)}, 2013, p. 2013.

\bibitem{maddern20171}
W.~Maddern, G.~Pascoe, C.~Linegar, and P.~Newman, ``1 year, 1000 km: The oxford
  robotcar dataset,'' \emph{The International Journal of Robotics Research},
  vol.~36, no.~1, pp. 3--15, 2017.

\bibitem{gaidon2016virtual}
A.~Gaidon, Q.~Wang, Y.~Cabon, and E.~Vig, ``Virtual worlds as proxy for
  multi-object tracking analysis,'' in \emph{Proceedings of the IEEE conference
  on computer vision and pattern recognition}, 2016, pp. 4340--4349.

\bibitem{arandjelovic2016netvlad}
R.~Arandjelovic, P.~Gronat, A.~Torii, T.~Pajdla, and J.~Sivic, ``Netvlad: Cnn
  architecture for weakly supervised place recognition,'' in \emph{Proceedings
  of the IEEE Conference on Computer Vision and Pattern Recognition}, 2016, pp.
  5297--5307.

\end{thebibliography}

\end{document}